\pdfoutput=1

\documentclass[11pt]{article}

\usepackage[]{acl}

\usepackage{times}
\usepackage{latexsym}

\usepackage[T1]{fontenc}

\usepackage[utf8]{inputenc}

\usepackage{amssymb}
\usepackage{microtype}
\usepackage{inconsolata}
\usepackage{float}
\usepackage{multirow}
\usepackage{tablefootnote}
\usepackage{graphicx}
\usepackage{amsmath}
\usepackage{booktabs}
\usepackage{caption}
\usepackage{subcaption}
\usepackage{bbold}
\usepackage{xspace}
\usepackage[capitalize,noabbrev]{cleveref}
\newcommand{\paperemoji}{\raisebox{-1.5pt}{\includegraphics[width=1.2em]{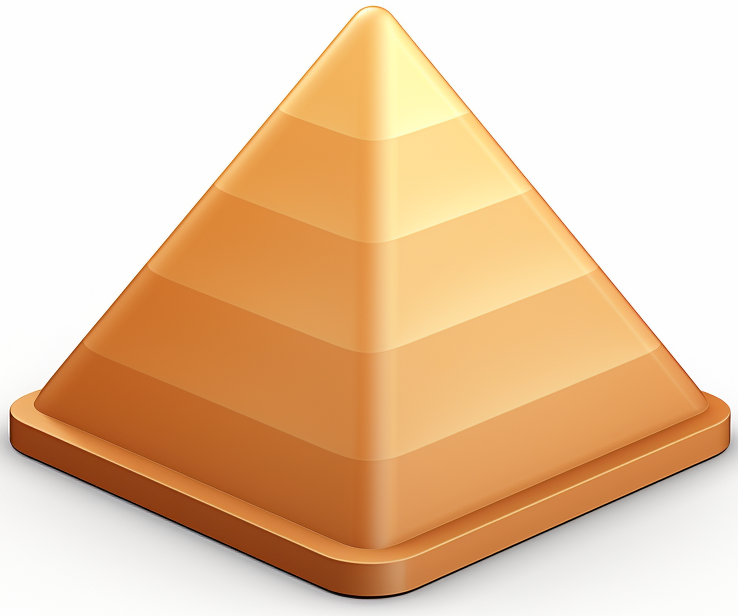}}}
\newcommand{\papertitle}{\textsc{AbsPyramid}\xspace}
\newcommand{\paperemojititle}{\paperemoji\papertitle}
\newcommand{\detectiondataname}{\textsc{AbsPyramid\textsubscript{[Det]}}\xspace}
\newcommand{\gendataname}{\textsc{AbsPyramid\textsubscript{[Gen]}}\xspace}
\newcommand{\nounrelation}{\textit{Noun-Entail}\xspace}
\newcommand{\verbrelation}{\textit{Verb-Entail}\xspace}
\newcommand{\eventrelation}{\textit{Event-Entail}\xspace}
\newcommand{\textitbf}[1]{\textit{\textbf{#1}}\xspace}
\newcommand{\textttbf}[1]{\texttt{\textbf{#1}}\xspace}
\usepackage{algorithm}
\usepackage{algpseudocode}

\definecolor{inputcolor}{HTML}{2ca02c}
\definecolor{outputcolor}{HTML}{1f77b4}
\definecolor{nouncolor}{HTML}{7030a0}
\definecolor{verbcolor}{HTML}{ef8600}
\definecolor{eventcolor}{HTML}{00717d}
\definecolor{instancecolor}{HTML}{2e75b6}
\definecolor{conceptcolor}{HTML}{c55a11}

\title{\paperemojititle: Benchmarking the Abstraction Ability of\\ Language Models with a Unified Entailment Graph}

\author{Zhaowei Wang$^1$, Haochen Shi$^1$, Weiqi Wang$^1$, \\\textbf{Tianqing Fang$^1$, Hongming Zhang$^2$, Sehyun Choi$^1$, Xin Liu$^1$, \& Yangqiu Song$^1$}\\
$^1$Department of Computer Science and Engineering, HKUST\\
$^2$Tencent AI Lab, Bellevue, USA\\
\texttt{\{zwanggy,wwangbw,tfangaa,schoiaj,xliucr,yqsong\}@cse.ust.hk} \\
\texttt{hshiah@connect.ust.hk,} \texttt{hongmzhang@global.tencent.com}}

\begin{document}
\maketitle
\begin{abstract}
Cognitive research indicates that abstraction ability is essential in human intelligence, which remains under-explored in language models. In this paper, we present \papertitle, a unified entailment graph of 221K textual descriptions of abstraction knowledge. While existing resources only touch nouns or verbs within simplified events or specific domains, \papertitle collects abstract knowledge for three components of diverse events to comprehensively evaluate the abstraction ability of language models in the open domain. Experimental results demonstrate that current LLMs face challenges comprehending abstraction knowledge in zero-shot and few-shot settings. By training on our rich abstraction knowledge, we find LLMs can acquire basic abstraction abilities and generalize to unseen events. In the meantime, we empirically show that our benchmark is comprehensive to enhance LLMs across two previous abstraction tasks\footnote{The code and data are available at \url{https://github.com/HKUST-KnowComp/AbsPyramid}.}.
\end{abstract}

\section{Introduction}
Abstraction is about finding common properties among different things and forming a broader concept, like the concept ``furniture'' subsuming ``sofa'' and ``table,'' a key dimension of human cognition~\cite{colung2003emergence, russell2010artificial}. With this ability, we can smoothly handle daily situations by learning from past experiences and generalizing to new circumstances~\cite{abstraction2013saitta}. Substantively, \citet{MINSKY1980117}, in his \textit{K-Theory}, suggested that our minds organize past experiences in a hierarchical pyramid, with higher parts corresponding to greater abstraction.

\begin{figure}[t]
    \centering
    \includegraphics[width=\columnwidth]{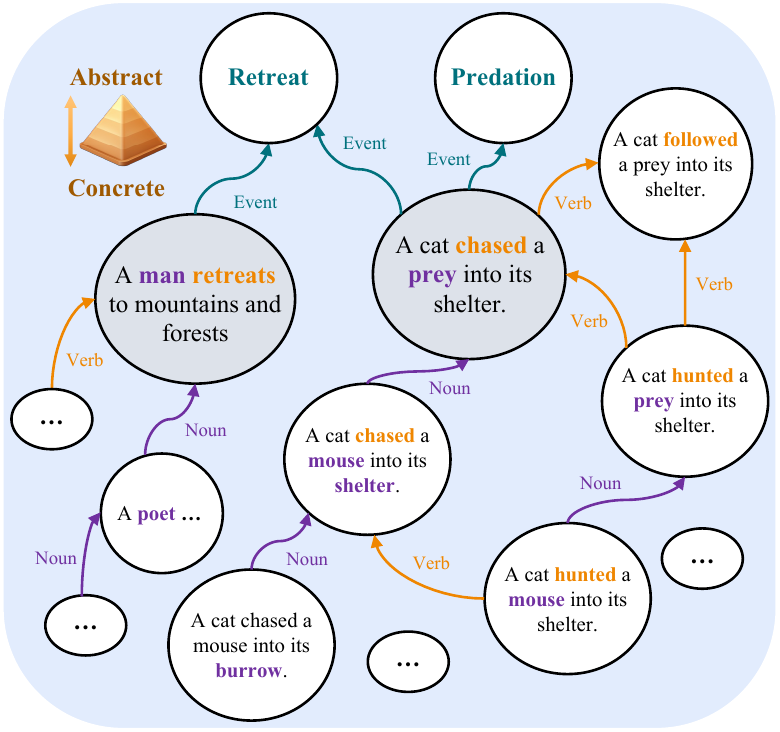}
    \caption{An illustration of our \papertitle benchmark. We identify three components of events (i.e., \textcolor{nouncolor}{\textitbf{Noun}}, \textcolor{verbcolor}{\textitbf{Verb}}, and \textcolor{eventcolor}{\textitbf{Event}} as a whole) and collect abstract concepts entailed by them.}
    \label{fig:intro_illustration}
\end{figure}

The NLP community has recently explored diverse, impressive abilities of LLMs, such as in-context learning~\cite{brown2020language}, multi-step reasoning~\cite{wei2022chain}, and instruction following~\cite{sanh2021multitask}. Meanwhile, the ability to abstract, a core dimension of human cognition, has received less attention in the studies of LLMs. Although sporadic works about abstraction knowledge exist, they focus solely on nouns or verbs within simplified events or specific domains, failing to consider a broader picture of abstraction.
One category of works is building an entailment graph of verbs, first proposed by \citet{berant2011global} with several techniques to enhance it in the following works~\cite{hosseini2018learning, mckenna2023smoothing}. Those works consider events as a \textit{verb} with two arguments (i.e., \textit{subject} and \textit{object}) and limit arguments to dozens of entity types to alleviate their graphs' sparsity issue. However, those simplifications considerably sacrifice the precise semantics of events. For example, the event ``a cat chased a mouse into its burrow'' in \cref{fig:intro_illustration} will be simplified into a tuple (animal, chase, animal), losing track of specific details of animals and location.
Other than verbs, \citet{he2022acquiring} annotated an abstraction dataset, AbstractATOMIC, about entities and events using the Probase taxonomy~\cite{wu2012probase}. While their work curated thousands of abstract concepts, it is limited to the social commonsense domain as base events are sampled from ATOMIC~\cite{sap2019atomic}.


Inspired by the cognitive study of \textbf{\textit{abs}}traction in the \textbf{\textit{pyramid}}-like hierarchy of human experiences~\cite{MINSKY1980117}, we present \papertitle, a unified entailment graph to comprehensively evaluate language models' abstraction ability. We curated abstract concepts entailed by each of the three components of an event\footnote{For readability, we use the term ``event'' in this paper. More accurately, our sampled data involve state, activity, and event, which can be summarized as a broader linguistic term: eventuality~\cite{mourelatos1978events, bach1986algebra}.}: nouns, verbs, and the event as a whole, unifying scopes and domains of all prior datasets. Specifically, we sample base events in textual descriptions from ASER~\cite{zhang2020aser, zhang2022aser}, an open-domain large-scale eventuality graph. We design heuristic rules to identify nouns and verbs from events and collect abstract concepts with WordNet~\cite{miller1995wordnet} and LLMs prompting. Those concept candidates are then crowdsourced for validity, resulting in a graph of 221K examples.
Compared with verb entailment graphs~\cite{berant2011global}, \papertitle retains specific and accurate semantics of base events. Our benchmark features a diverse array of syntactic roles for real arguments instead of relying on (\textit{subject}, \textit{verb}, \textit{object}) tuples with entity types. In contrast to AbstractATOMIC~\cite{he2022acquiring}, our benchmark covers abstraction knowledge beyond the social commonsense thanks to the open domain corpora used in ASER. Also, we use LLMs to broaden collected abstract concepts, complementing the coverage of taxonomies.

On the \papertitle benchmark, we investigate whether LLMs can (1) identify valid abstract concepts and (2) generate abstract concepts.
The evaluation results on 26 popular language models reveal that: (1) LLMs encounter difficulties understanding abstraction knowledge under both zero-shot and in-context learning settings. (2) In contrast, fine-tuned language models perform better at comprehending abstraction knowledge, especially for nouns. (3) Our benchmark incorporates comprehensive abstraction knowledge, which can improve LLMs' performance significantly across verb entailment graphs and AbstractATOMIC.
To the best of our knowledge, \papertitle presents the first comprehensive evaluation of LLMs' abstraction ability. Our benchmark and experiment results provide valuable insights into the abstraction ability of language models and the progress of artificial intelligence within LLM.

\section{Related Work}
While the NLP community has studied various abilities of LLMs~\cite{wei2022emergent,chowdhery2022palm,ouyang2022training,chung2022scaling,zhou2022least}, the abstraction ability of LLMs remains insufficiently studied. Unlike existing works that focus on entity-level abstraction~\cite{clark2000exploiting,van2009deriving,song2011short,song2015open,gong2016representing}, our research delves into event-level abstraction with only a few works investigating some restricted aspects:

\paragraph{Verb Entailment Graph:} \citet{berant2011global} first proposed the task of entailment graph construction of verbs. Following their work, various methods have been proposed to build better verb entailment graphs~\cite{hosseini2018learning,hosseini2019duality,hosseini2021open,guillou2020incorporating,chen2022entailment,tianyi2022cross,mckenna2021multivalent,mckenna2023smoothing}. Nonetheless, those works consider verbs as binary relations with two arguments from a small set of entity types (e.g., 49 types in FIGER~\cite{hosseini2018learning}), distorting the original semantics.

\paragraph{AbstractATOMIC:} \citet{he2022acquiring} presented an annotated abstraction dataset. They recognized entities in head events from ATOMIC~\cite{sap2019atomic} and crowdsourced abstract concepts from the Probase taxonomy~\cite{wu2012probase} for recognized entities and head events. Even though they compiled a dataset comprising thousands of examples, it is specific to the social commonsense domain due to the base events sampled from ATOMIC.

\paragraph{Textual and Linguistic Entailment:} Besides the entailment between verbs, recognizing textual entailment has long been a vital task in 
the realm of NLP~\cite{cooper1996using,dagan2005PASCAL}, also known as \textit{natural language inference} (NLI). Researchers have built many large-scale datasets of NLI~\cite{conneau-etal-2018-xnli,marilyn2018broad,yixin2020adversarial} and its variants~\cite{ alex2019glue,bhavana2021explaining,chen-etal-2023-propsegment}. 

While similar to our task, textual entailment employs a relaxed definition of whether a human reader would \textitbf{typically infer} a \textit{hypothesis} from a given \textit{premise}~\cite{maccartney2007natural,daniel2018defining} instead of abstraction of the \textit{premise}. For example, in SNLI~\cite{samuel2015large}, we can infer \textit{a boy is holding his arms out} from the premise \textit{a boy looks down and spreads his arms wide} without any abstraction involved. 
In contrast, our work follows the definition of linguistic entailment~\cite{beth1955semantic}, which arises from the semantics of linguistic expressions and is enforced by lexical meanings plus the laws of logic~\cite{murphy2010lexical,houndmills2007presupposition}. For instance, \textit{Max is a playful puppy} entails \textit{Max is a dog} since one cannot be a playful puppy without being a dog.


\begin{figure}[t]
    \centering
    \includegraphics[width=0.95\columnwidth]{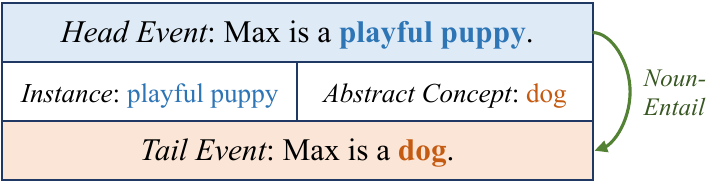}
    \caption{An illustration of the structure of abstraction knowledge, where \textitbf{entailment relation} is \nounrelation.}
    \label{fig:structure_example}
\end{figure}
\section{Abstraction Knowledge Structure}
\label{sec:knowledge_structure}
\papertitle represents a large-scale abstraction repository of events in textual descriptions. This unified entailment graph contains 221K 
five-element tuples with the format of (\textitbf{head event}, \textitbf{entailment relation}, \textitbf{tail event}, \textitbf{instance}, \textitbf{abstract concept}). In each tuple, we identify an \textitbf{instance} in the \textitbf{head event} and collect an \textitbf{abstract concept} for it. Particularly, instances are identified from three components of the head event: nouns, verbs, and head event as a whole. Then, we replace the instance with its abstract concept to construct the \textitbf{tail event}, resulting in the tail event being linguistically entailed by the head event. According to three kinds of instances, we define three types of \emph{\textbf{entailment relation}}: \nounrelation, \verbrelation, and \eventrelation. We elaborate on each tuple element with a concrete example in \cref{fig:structure_example}.

\section{Data Curation Pipeline}
To build \papertitle, we create a crowdsourcing framework
that allows for a scalable, broad collection of abstraction knowledge in the abovementioned format.

\subsection{Compiling Head Events}
We randomly sample 17K base eventualities from ASER as head events. Since ASER is an automatically extracted graph, some noisy extraction results may affect the quality of our benchmark. Thus, we design elaborate rules to clean ASER using lexical and dependency parsing features (Details in \cref{app:aser_cleaning}). Meanwhile, ASER is extracted from six open domain corpora spanning Wikipedia\footnote{https://dumps.wikimedia.org/enwiki}, NYT~\cite{sandhaus2008new}, Yelp\footnote{https://www.yelp.com/dataset/challenge}, Reddit\footnote{https://www.reddit.com/r/datasets/comments/3bxlg7}, etc. We only sample eventualities from NYT and Wikipedia due to the less formal nature of other corpora, such as diverse styles of comments on Yelp. To collect more general events, we replace tokens referring to people with a \texttt{Person} variable (e.g., replace I/we/she/... with \texttt{PersonX/Y/Z}), following previous work~\cite{sap2019atomic}.

\subsection{Identifying Instances}
As mentioned earlier, our benchmark defines three entailment relations. For \eventrelation, we can directly use head events as identified instances. More intricately, we need to identify nouns and verbs as instances within head events when dealing with \nounrelation and \verbrelation. We design an algorithm to heuristically match nouns and verbs based on parsing results (e.g., POS-tags) provided by ASER (Details in \cref{app:matching_nouns_and_verbs}).

\subsection{Collecting Abstract Concepts}
Then, we collect abstract concepts for those identified instances through two methods: (1) retrieving from non-contextualized taxonomy and (2) prompting LLMs to generate candidates in free form.

\paragraph{Pilot Study:} There are two taxonomies of words containing abstract concepts: WordNet~\cite{miller1995wordnet} and Probase~\cite{wu2012probase}. WordNet contains hypernym relations, words with a broad meaning that more specific words (i.e., hyponyms) fall under. Probase automatically extracts instance-concept relations of nouns from corpora. Both aggregate all senses of each word without context.

Our pilot study reveals that WordNet effectively covers more than 90\% of verbs within head events. Nonetheless, the coverage of nouns is unsatisfactory, as we can build a gigantic space of nominal phrases by adding modifiers. For example, we can easily form numerous phrases of ``dog'' by adding ``guard,'' ``hunting,'' or ``white,'' etc. Our pilot study finds that only 6.3\% of nominal phrases in head events are covered by WordNet. Likewise, the coverage of Probase is also unacceptable (29.6\%).

\paragraph{Abstract Concepts for Nouns:}
Due to the limited coverage of nouns in taxonomies, we collect hypernyms for nouns by prompting an LLM. In detail, we prompt ChatGPT under the in-context learning setting with the standard task-instruction-then-exemplar prompts~\cite{west2022symbolic}: 
\begin{center}
\resizebox{0.9\linewidth}{!}{
\begin{tabular}{l}
  \textbf{\texttt{ <INSTRUCTION> }} \\
  \textbf{\texttt{ \textcolor{inputcolor}{<EX$_1$-IN>}\textcolor{outputcolor}{<EX$_1^{(1)}$-OUT>} \ldots \textcolor{outputcolor}{<EX$_1^{(K)}$-OUT>} }}
  \\
  \ldots \\
    \textbf{\texttt{ \textcolor{inputcolor}{<EX$_{N}$-IN>}\textcolor{outputcolor}{<EX$_{N}^{(1)}$-OUT>} \ldots \textcolor{outputcolor}{<EX$_{N}^{(K)}$-OUT>} }}\\
    \textbf{\texttt{ \textcolor{inputcolor}{<EX$_{N+1}$-IN>} }}
\end{tabular}
} 
\end{center}

\noindent where \textttbf{<INSTRUCTION>} describes the task of finding abstract concepts of a noun in our case. The input \textttbf{<EX$_{i}$-IN>} is a head event with an identified noun, with output \textttbf{<EX$_{i}^{(k)}$-OUT>} being an abstract concept. Given such a prompt, ChatGPT compactly generates $K$ abstract concepts for each testing input. In the meantime, we design another prompt to elicit challenging negative examples that are highly related but not abstract concepts, such as ``stream course'' for ``stream'' in ``the stream creates a peaceful ambiance.'' Prompts are shown in \cref{app:prompts_abstract_concept} concretely, with $N$ and $K$ equal to 10.


\paragraph{Abstract Concepts for Verbs:} We collect abstract concepts for verbs using hypernyms from WordNet, as verbs are well covered. We link verbs into WordNet and employ GlossBERT~\cite{huang-etal-2019-glossbert}, a word-sense disambiguation (WSD) model, to select each verb's correct (at least most probable) word sense. Then, hypernyms of the correct word sense are collected as abstract concepts. 


\paragraph{Abstract Concepts for Events:} Events are more complex than nouns and verbs without relevant taxonomy. Thus, we again prompt ChatGPT to collect phrasal abstract concepts of each head event. We use the prompts similar to nouns with slight changes in verbalizing input tuples (More details in \cref{app:prompts_abstract_concept}). $N$ and $K$ are equal to 10.

\subsection{Dataset Annotation}
The last step of our data curation pipeline is to verify the validity of automatically collected abstract concepts. We create an annotation task for each entailment relation on Amazon Mechanical Turk (MTurk). In those tasks, we first give annotators detailed instructions about the validity of abstract concepts, like explanations of hypernyms. We provide annotators with five-element tuples, as mentioned in \cref{sec:knowledge_structure}, asking them whether each abstract concept is valid. For \verbrelation, we also provided meanings of each verb from WordNet for better understanding. Meanwhile, to ensure annotation quality, we introduce two qualification tests and two rounds of annotation refinement. Details of quality control and annotation agreements are shown in \cref{app:annotation_details}.


\section{\paperemojititle Overview}
In this section, we carry out a thorough analysis of our benchmark \papertitle.
\begin{table}[t]
    \small
    \setlength{\tabcolsep}{4.8pt}
    \centering
	\begin{tabular}{l|ccccc}
	    \toprule
        \textbf{\textsc{Rel.}}&\textbf{\# Total}&\textbf{\# Train}&\textbf{\# Valid}&\textbf{\# Test}&\textbf{\% Pos}\\
		\midrule
		\textsc{Noun}& 98,783 & 79,034 & 9,874 & 9,875 & 58.98 \\
		  \textsc{Verb}& 59,542 & 47,669 & 5,939 & 5,934 & 52.29 \\
            \textsc{Event}& 62,472 & 49,988 & 6,237 & 6,247 & 64.77 \\
            \textsc{All}& 220,797 & 176,691 & 22,050 & 22,056 & 58.82\\
		\bottomrule
	\end{tabular}
	\caption{Statistics of \papertitle. \textbf{Pos} denotes positive rates. \textbf{\textsc{Rel.}} indicates entailment relations. We split data into training, validation, and test sets (80:10:10).}
	\label{table:dis_stat}
\end{table}

\subsection{Benchmark Statistics}
\papertitle is a large-scale benchmark comprising about 221K abstraction examples. Specific details are shown in \cref{table:dis_stat}. For breakdown details, we collected more than 98K, 59K, and 62K tuples for \nounrelation, \verbrelation, and \eventrelation. To better understand our benchmark, We compare it with the Levy/Holt dataset~\cite{levy2016annotating, holt2018probabilistic}, a dataset heavily used to evaluate verb entailment graphs, and AbstractATOMIC~\cite{he2022acquiring}. Four statistical metrics are computed for multi-dimensional comparison, including data size, vocabulary size, percentage of unique abstract concepts, and social domain proportions, with results as follows. 

Previous studies show that content generated by LMs, ChatGPT in our case, might lack diversity~\cite{welleck2019neural}. From \cref{tab:stat_compare}, we can find that our benchmark has a much larger \textbf{data size} and \textbf{vocabulary size} than previous resources, showing the lexical diversity of our benchmark. In particular, the vocabulary size is more than three times that of prior resources.

We also compute the \textbf{percentage of unique abstract concepts} based on BLEU soft uniqueness~\cite{zhu2018texygen,west2022symbolic}. An abstract concept $x$ is unique if $BLEU_1(C, x) \leq 0.5$, where $C$ is all concepts that share the same head event and identified instance with $x$, and 0.5 is an empirical threshold. Our benchmark has a percentage on par with other datasets, showing the efficacy of our data curation pipeline. Last, we also report the \textbf{social domain proportions}, where we count head events with \texttt{Person} variables. As shown in \cref{tab:stat_compare}, all head events in AbstractATOMIC contain \texttt{Person} variables since they are sampled from ATOMIC. In contrast, 32.19\% of head events in \papertitle pertain to daily life experiences.

\begin{table}[t]
    \centering
    \small
    \setlength{\tabcolsep}{4.5pt}
    \begin{tabular}{l|cccc}
        \toprule 
        \textbf{Dataset} & \textbf{Data (K)} & \textbf{Vocab. (K)} & \textbf{Unique} & \textbf{Social} \\ 
        \midrule
        \textsc{Noun} & 98.78 & 20.95 & 93.57 & 19.88 \\
        \textsc{Verb} & 59.54 & 11.86 & 95.74 & 40.02 \\
        \textsc{Event} & 62.47 & 19.04 & 73.43 & 36.15 \\
        \textsc{All} & 220.80 & 29.42 & 88.26 & 32.19 \\
        \midrule
        AbsAtomic & 92.23 & 8.99 & 89.42 & 100.00 \\
        Levy/Holt & 18.41 & 5.62 & 87.85 & 38.17 \\  
        \bottomrule
    \end{tabular}
    \caption{Dataset comparison. Data size, vocabulary size, percentage of unique abstract concepts, and social domain proportion are listed.} 
    \label{tab:stat_compare}		
\end{table}

\subsection{Evaluation Tasks}
We study two tasks on our benchmark, abstraction detection and generation, to evaluate whether LLMs can detect and generate abstraction knowledge. In the detection task, models are given a five-element tuple (in \cref{sec:knowledge_structure}) and are asked to decide if the abstract concept is valid. We split collected abstraction knowledge into training, validation, and test sets (80:10:10) to form the \detectiondataname dataset (in \cref{table:dis_stat}). In the generation task, models are requested to generate abstract concepts for a given tuple. We remove tuples with invalid abstract concepts and form \gendataname dataset in~\cref{table:gen_stat}. We ensure that tuples sharing the same head event and identified instances are in the same set for both datasets.

\section{Abstraction Detection Experiment}
In this section, we conduct extensive experiments on the \detectiondataname dataset to evaluate an abundance of language models and provide comprehensive analyses.

\subsection{Experiment Setup}
\paragraph{Evaluation Metric:}
We calculate Accuracy, Macro F1-score, and ROC-AUC between predicted and ground-truth labels to evaluate all models.

\paragraph{Models} 
We evaluate four categories of LMs. \textbf{(1) PLM + FT:} We fine-tune pre-trained LMs: BERT~\cite{devlin2018bert}, RoBERTa~\cite{liu2019roberta}, and DeBERTa~\cite{he2020deberta}, in the base and large sizes. \textbf{(2) NLI + Zero\&FT:} We include four models fine-tuned on NLI data: BART-large-mnli~\cite{lewis2020bart}, RoBERTa-base/large-mnli~\cite{liu2019roberta}, and DeBERTa-large-mnli~\cite{he2020deberta}. We assess the zero-shot capability of those models and fine-tune them on our dataset. \textbf{(3) LLM + LoRA:} We fine-tune representative LLMs with LoRA~\cite{hu2021lora}: Llama2 (7B, 13B) and Llama2-Chat (7B, 13B)~\cite{touvron2023llama}, Falcon (7B) and Falcon-Instruct (7B)~\cite{penedo2023refinedweb}, and Mistral (7B) and Mistral-Instruct (7B)~\cite{jiang2023mistral}. \textbf{(4) LLM API:} We assess a series of closed-source LLMs under the zero-shot and in-context learning setups, covering GPT3.5~\cite{ouyang2022training}, ChatGPT~\cite{openai2023chatgpt}, and GPT4~\cite{openai2023gpt4}. We use a standard and a CoT prompt~\cite{kojima2022large}. See implementation details in \cref{app:implementation_details}.

\begin{table}[t]
    \small
    \setlength{\tabcolsep}{4.5pt}
    \centering
	\begin{tabular}{l|cccccc}
	    \toprule
        \textbf{\textsc{Rel.}}&\textbf{\# Total}&\textbf{\# Train}&\textbf{\# Valid}&\textbf{\# Test}&\textbf{Avg-Ref}\\
		\midrule
		\textsc{Noun}& 58,266 & 52,440 & 2,910 & 2,916 & 5.58\\
		  \textsc{Verb}& 31,132 & 28,018 & 1,556 & 1,558 & 2.90\\
            \textsc{Event}& 40,466 & 36,446 & 2,006 & 2,014 & 4.57\\
            \textsc{All}& 129,864 & 116,904 & 6,472 & 6,488 & 4.33\\
		\bottomrule
	\end{tabular}
	\caption{The statistics of generation data. \textbf{Avg-Ref} means the average references per identified instance. \textbf{\textsc{Rel.}} stands for entailment relations. Tuples are split into training, validation, and test sets (90:5:5).}
	\label{table:gen_stat}
\end{table}

\begin{table*}[t]
    \small
    \setlength{\tabcolsep}{4.9pt}
	\centering
	\begin{tabular}{l|l||ccc|ccc|ccc}
	\toprule
        \multirow{2}{*}{\textbf{Methods}}&\multirow{2}{*}{\textbf{Backbone}}&\multicolumn{3}{c|}{\textbf{Noun}} &\multicolumn{3}{c|}{\textbf{Verb}}&\multicolumn{3}{c}{\textbf{Event}}\\
	&&\textbf{Acc}&\textbf{Ma-F1} &\textbf{AUC} &\textbf{Acc}&\textbf{Ma-F1} &\textbf{AUC} & \textbf{Acc}&\textbf{Ma-F1}&\textbf{AUC} \\
            \midrule
            \textbf{Random} & \multicolumn{1}{|c||}{-} &50.00 & 49.56 & 50.00 & 50.00 & 49.95 & 50.00 & 50.00 & 48.98 & 50.00\\
            \textbf{Majority Vote} & \multicolumn{1}{|c||}{-} &59.30 & - & 50.00 & 53.15 & - & 50.00 & 64.14 & - & 50.00\\
		\midrule
            \multirow{4}{*}{\textbf{NLI + Zero}}&BART-large-mnli&71.24&68.13&75.67&56.25&47.17&62.33&70.69&65.81&69.33\\
            &RoBERTa-large-mnli&68.66&63.18&75.42&55.73&45.54&61.27&70.47&63.07&68.60\\
            &DeBERTa-base-mnli&68.77&65.81&72.79&56.42&48.08&61.55&66.30&62.88&66.40\\
            &DeBERTa-large-mnli&73.18&71.08&78.12&56.93&49.28&63.16&66.82&64.03&68.27\\
            \midrule
            \multirow{4}{*}{\textbf{NLI + FT}}&BART-large-mnli&85.75&85.12&90.80&64.96&64.96&68.60&74.61&69.75&77.71\\
            &RoBERTa-large-mnli&86.15&85.34&90.87&64.61&64.26&69.46&76.88&70.73&77.94\\
            &DeBERTa-base-mnli&85.59&84.61&90.43&65.50&65.47&\underline{69.87}&76.98&70.12&77.90\\
            &DeBERTa-large-mnli&86.62&85.83&91.00&\textbf{66.04}&\textbf{65.96}&\textbf{70.51}&76.48&69.96&77.42\\
            \midrule
            \multirow{6}{*}{\textbf{PLM + FT}}&BERT-base&85.09&84.14&89.94&64.26&64.20&68.06&76.45&69.94&78.22\\
            &BERT-large&85.94&85.12&90.37&63.58&63.58&68.03&75.27&69.61&77.57\\
            &RoBERTa-base&84.23&83.25&89.58&63.55&63.53&68.12&76.53&70.41&77.62\\
            &RoBERTa-large&85.27&84.44&90.59&64.98&64.98&69.23&77.09&70.56&78.07\\
            &DeBERTa-base&84.09&83.03&89.74&63.50&63.45&68.03&75.75&69.57&77.30\\
            &DeBERTa-large&86.89&86.11&90.98&\underline{65.54}&\underline{65.52}&69.11&76.69&70.31&78.06\\
            \midrule
            \multirow{8}{*}{\textbf{LLM + LoRA}}            &Falcon (7B)&87.06&86.36&91.42&63.92&63.79&68.06&75.83&70.51&77.77\\
            &Falcon-Ins (7B)&86.04&85.43&91.10&64.00&63.96&68.53&76.50&70.72&77.50\\
            &Mistral (7B)&87.62&87.05&91.53&65.08&64.66&69.58&\underline{77.24}&70.57&77.97\\
            &Mistral-Ins (7B)&87.59&86.99&91.42&64.81&64.78&69.51&77.22&70.69&78.52\\
            &Llama2 (7B)&87.56&86.82&91.52&65.07&64.79&69.27&76.45&70.53&78.28\\
            &Llama2-Chat (7B)&86.71&86.17&91.79&64.96&64.54&68.95&76.80&70.15&77.92\\
            &Llama2 (13B) &\underline{88.03}&\underline{87.40}&\textbf{92.31}&65.13&64.64&69.50&76.87&\textbf{70.83}&\textbf{79.34}\\
            &Llama2-Chat (13B)&\textbf{88.20}&\textbf{87.49}&\underline{92.05}&65.07&65.00&69.74&\textbf{77.27}&\underline{70.82}&\underline{78.60}\\
            \midrule
            \midrule
            \multirow{6}{*}{\textbf{LLM API}}
            &GPT 4 &80.50&78.70&-&56.30&53.84&-&71.30&66.89&-\\
            &GPT 3.5&67.00&62.45&-&56.30&55.90&-&65.60&58.23&-\\
            &ChatGPT&74.00&72.27&-&56.30&55.71&-&68.20&63.22&-\\
            &ChatGPT (CoT) &62.90&62.88&-&56.20&53.89&-&67.30&61.47&-\\
            &ChatGPT (10-shot ICL)&76.10&74.60&-&58.60&58.51&-&68.90&60.51&-\\
            &ChatGPT (CoT + 10-shot)&75.40&74.08&-&59.20&58.91&-&68.20&62.70&-\\
		\bottomrule
	\end{tabular}
	\caption{Performance on the test set of \detectiondataname. We trained models on three entailment relations separately. We bold the best score and underline the second-best score. Acc, Ma-F1, and AUC denote Accuracy, Macro F1-score, and ROC-AUC. See the performance on the validation set in \cref{app:validation_results}.}
    \label{tab:main_eval_test}
\end{table*}

\subsection{Main Evaluation}
We train LMs on each entailment relation separately and present results on \detectiondataname in \cref{tab:main_eval_test}. We observe that fine-tuned LMs can detect abstraction knowledge of \nounrelation with impressive performance. For example, Llama2-Chat (13B) correctly classifies 88.20\% of the test data. Meanwhile, models struggle to achieve similar scores on \verbrelation relation. The difficulty of \verbrelation might come from the diversity of word senses we collected from WordNet.

NLI models show some zero-shot ability, especially on \nounrelation and \eventrelation. For instance, DeBERTa-large-mnli achieves an accuracy of 73.18\% on \nounrelation higher than that of ``random'' and ``majority vote.'' This finding might be due to some similarity between NLI and our task. Moreover, fine-tuning NLI models cannot improve performance compared with LMs in \textbf{PLM + FT}.

Besides, fine-tuned LLMs can obtain scores comparable to or even higher than fully fine-tuned models, whilst we only tuned 0.3-0.5\% parameters with LoRA. The performance only improves marginally when we increase the parameters, such as Llama2 (7B) to Llama2 (13B). Meanwhile, the instruction-tuned counterparts cannot lead to distinct increases but some fluctuations as they learned more about the instruction following and conversations, which are irrelevant to our task.

\begin{table*}[t!]
\setlength{\tabcolsep}{4pt}
\centering
\small
\begin{tabular}{l||ccc|ccc|ccc|ccc}
	\toprule
        \multirow{2}{*}{\textbf{LLM + LoRA}}&\multicolumn{3}{c|}{\textbf{Noun}} &\multicolumn{3}{c|}{\textbf{Verb}}&\multicolumn{3}{c|}{\textbf{Event}}&\multicolumn{3}{c}{\textbf{All}}\\
	&\textbf{Acc}&\textbf{Ma-F1} &\textbf{AUC} &\textbf{Acc}&\textbf{Ma-F1} &\textbf{AUC}&\textbf{Acc}&\textbf{Ma-F1} &\textbf{AUC} &\textbf{Acc}&\textbf{Ma-F1} &\textbf{AUC}  \\
		\midrule
            Falcon (7B) &87.11&86.31&91.26&64.68&64.34&69.50&76.55&70.47&78.52&78.15&76.53&84.78\\
            Falcon-Ins (7B) &87.07&86.30&90.91&64.71&64.70&69.16&\textbf{77.22}&70.95&78.26&78.28&76.92&84.64\\
            Mistral (7B) &87.77&87.01&91.68&\textbf{65.96}&\underline{65.60}&\textbf{70.34}&76.61&70.91&78.88&78.71&77.15&85.40\\
            Mistral-Ins (7B) &87.80&\underline{87.09}&91.47&65.44&65.35&69.94&77.08&71.08&\textbf{79.50}&\underline{78.75}&\underline{77.37}&85.38\\
            Llama2 (7B)&\underline{87.92}&87.09&\textbf{91.80}&64.95&64.47&69.59&\underline{77.16}&71.05&78.75&78.69&76.95&85.39\\
            Llama2-Chat (7B)&87.56&86.79&\underline{91.79}&64.11&63.98&69.48&76.55&70.53&77.84&78.09&76.98&85.00\\
            Llama2 (13B)&\textbf{88.02}&\textbf{87.41}&91.73&\underline{65.84}&\textbf{65.84}&\underline{70.16}&77.11&\underline{71.13}&78.93&\textbf{78.99}&\textbf{77.83}&\textbf{85.73}\\
            Llama2-Chat (13B)&87.76&87.00&91.59&65.08&64.87&70.02&76.98&\textbf{71.16}&\underline{79.39}&78.67&77.17&\underline{85.49}\\
		\bottomrule
  \end{tabular}
\caption{The performance of LLMs on the test set of \detectiondataname under the multi-relation setting. We bold the best score and underline the second-best score. See \cref{app:validation_results} for performance on validation sets.}
\label{tab:multi_relation_test}
\end{table*}

\begin{table}[t]
    \small
    \centering
	\begin{tabular}{l||cccc}
	    \toprule
		\textbf{Models}&\textbf{Acc}&\textbf{Ma-F1}&\textbf{AUC}&\textbf{APS}\\
		\midrule
            \multirow{1}{*}{Aug MC} & - & - & - & 18.70 \\
            CNCE MC & - & - & - & 19.50 \\
            EGT2 & - & - & - & 31.90 \\
            \midrule
            Falcon (7B) &67.55&63.82&80.06&\multicolumn{1}{l}{39.97\textsubscript{$\uparrow$8.07}}\\
            Mistral (7B) &79.32&\textbf{72.66}&\textbf{81.42}&\multicolumn{1}{l}{\textbf{53.25}\textsubscript{$\uparrow$\textbf{21.35}}}\\
            Llama2 (7B) &78.69&71.07&79.51&\multicolumn{1}{l}{44.25\textsubscript{$\uparrow$12.35}}\\
            Llama2 (13B) &\textbf{82.11}&71.25&79.84&\multicolumn{1}{l}{45.11\textsubscript{$\uparrow$13.21}}\\
		\bottomrule
	\end{tabular}
	\caption{Zero-shot performance on Levy/Holt dataset with LLMs fine-tuned on our dataset. \textbf{APS} is average precision score when precision $> 0.5$ and shows improvements compared with EGT2.}
	\label{tab:zero_shot_levy}
\end{table}

\begin{figure}[t]
    \centering
    \includegraphics[width=0.95\columnwidth]{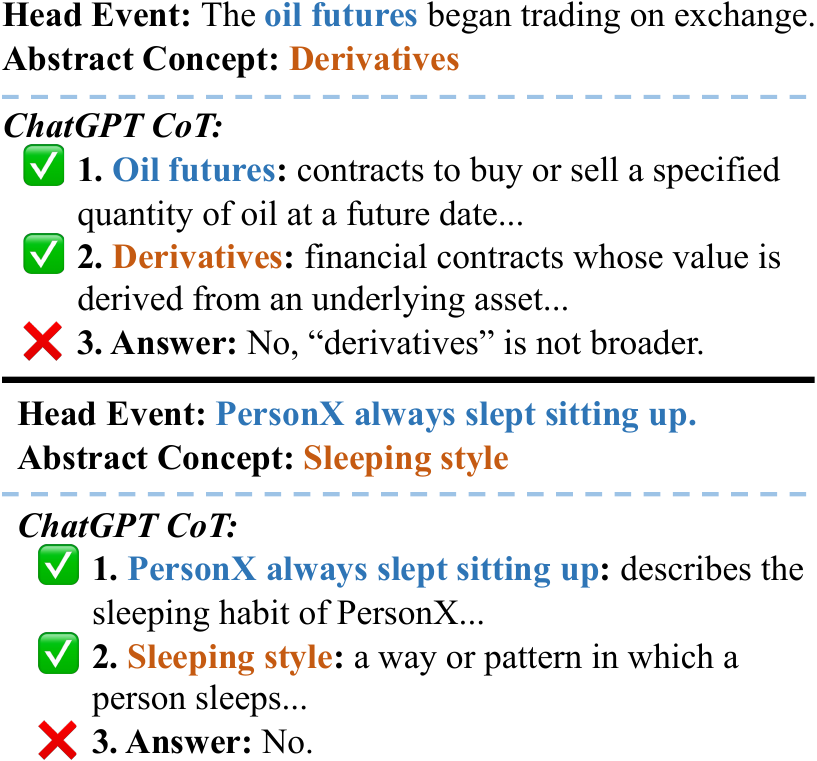}
    \caption{Error Analysis. We find hallucinations within zero-shot CoT of ChatGPT with correct explanations but wrong conclusions.}
    \label{fig:error_analysis}
\end{figure}

\subsection{Analysis of ChatGPT Series Models}
We can see that ChatGPT and GPT3.5 obtain acceptable performance on \detectiondataname in the zero-shot scenario (\cref{tab:main_eval_test}), such as accuracy scores of 74.00\% and 67.00\% on \emph{Noun-Entail}. However, the ChatGPT series models still lag behind fine-tuned LMs by a large margin, although GPT4 performs better than ChatGPT. Meanwhile, we tested the performance of ChatGPT with ten exemplars under the in-context learning setup, denoted as ``ChatGPT (10-shot ICL).'' With exemplars, the scores of ChatGPT are raised by 2-3 points but not a substantial improvement since the answer format (i.e., ``Yes'' or ``No'') is simple to understand without exemplars.

To explore if the ChatGPT can explain its own decisions, we examine ChatGPT with zero-shot chain-of-thought prompting signified as ``ChatGPT (CoT),'' where it is asked to explain given words first and then give the answer. Each metric exhibits varying levels of decline, with particular emphasis on \emph{Noun-Entail}. This indicates that ChatGPT cannot explain and provide an answer simultaneously. We conduct an error analysis, as illustrated in \cref{fig:error_analysis}, to unravel why. The examples show that ChatGPT can explain the meanings of given words but yields hallucinations~\cite{ji2023survey,huang2023survey} when concluding. 
We discover that providing a few exemplars can assist, indicated as ``ChatGPT (CoT + 10-shot)'' in \cref{tab:main_eval_test}. We present all prompts and verify the robustness of zero-shot and CoT prompts in \cref{app:chatgpt_prompt_robustness}.

\subsection{Multi-Relation Learning}
\label{sec:mutli_relation_learning}
While prior experiments treated each relation separately, we train all entailment relations jointly in this section. The results in \cref{tab:multi_relation_test} show that LLMs can learn abstraction knowledge of multiple relations, with performance comparable to that of training on each relation separately (\cref{tab:main_eval_test}). Generally, Llama2 (13B) performs best on the merged test set, while varying models get higher performance on each entailment relation. Comparing Llama2 (7B) with Llama2 (13B), we again affirm that scaling up models only leads to marginal improvements.

\subsection{Transferring to Other Sources}
This section investigates whether the abstraction knowledge from our benchmark can be transferred to other tasks that require the abstraction knowledge~\cite{berant2011global,he2022acquiring}.

\paragraph{Verb Entailment Graph:} In this task, we evaluate models on the primarily used Levy/Holt dataset~\cite{levy2016annotating, holt2018probabilistic}, whose statistics are shown in \cref{tab:stat_compare}. We directly experiment with the LLMs fine-tuned on our data (under the multi-relation setting in \cref{sec:mutli_relation_learning}) to test the zero-shot transferring ability. Following previous works~\cite{hosseini2021open}, we also compute the metric ``average precision score'' when precision is higher than 50\%. As shown in \cref{tab:zero_shot_levy}, LLMs fine-tuned on our dataset surpass previous works a lot, including Aug MC~\cite{hosseini2018learning}, CNCE MC~\cite{hosseini2019duality}, and EGT2~\cite{chen2022entailment}. For example, Mistral (7B) achieves the best APS of 53.25, higher than the strongest baseline, EGT2, by over 20 points. For a complete comparison, we also test instruction-tuned LLMs as another baseline in \cref{app:result_transferring}.

\begin{figure}[t]
    \centering
    \includegraphics[width=\columnwidth]{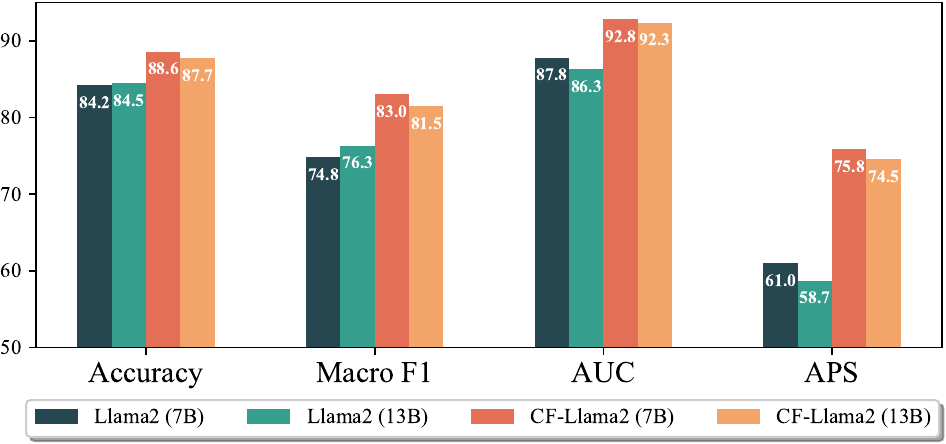}
    \caption{The fine-tuning performance on the Levy/Holt dataset. \textbf{CF} stands for continually fine-tuning.}
    \label{fig:finetune_levy_holt}
\end{figure}

We further test whether knowledge can be transferred in the fine-tuning setup. We continually fine-tune with LoRA LLMs that are first trained on our dataset. They are compared with LLMs fine-tuned from pre-trained configurations. Since the Levy/Holt dataset does not own a training set, we treat the validation set as the training set and do not tune hyperparameters. From \cref{fig:finetune_levy_holt}, the results show that training on our benchmark significantly boosts the performance of LLMs on all metrics. Particularly, the average precision score of Llama2 (7B) rises from 61.0 to 75.8 if we first fine-tune it on our benchmark. These experiments demonstrate that our benchmark is comprehensive to boost performance in both zero-shot and fine-tuning setups.

\begin{figure}[t]
    \centering
    \includegraphics[width=\columnwidth]{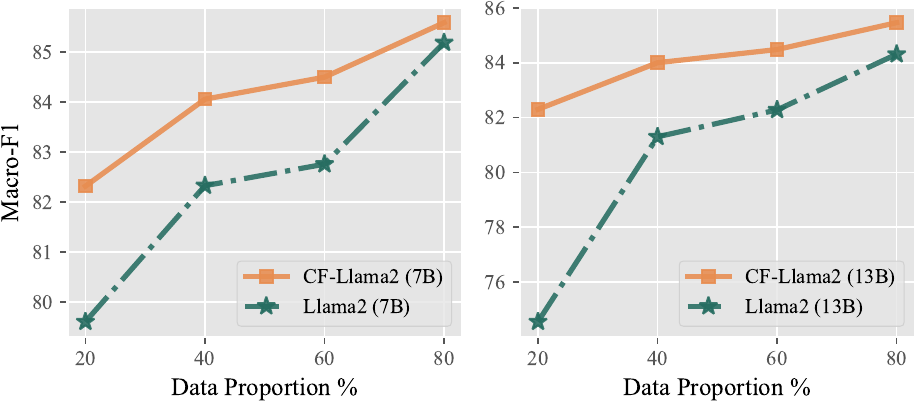}
    \caption{Few-shot performance on AbstractATOMIC. \textbf{CF} stands for continually fine-tuning.}
    \label{fig:abstract_atomic}
\end{figure}

\paragraph{AbstractATOMIC}
To further verify the comprehensiveness of our benchmark, we fine-tuned LLMs under the few-shot setting on the AbstractATOMIC dataset, where we start from 20\% of training data and increase the proportion by 20\% each time. Similarly, we fine-tuned two categories of LLMs: pre-trained models and models initially trained on our dataset. While only a modest fraction of our dataset falls under the social domain (in \cref{tab:stat_compare}), we discover that our dataset still can significantly enhance performance on AbstractATOMIC, as displayed in \cref{fig:abstract_atomic}. The results show that our dataset contains comprehensive abstract knowledge, which can help models generalize to a specific domain. We include full results of more LLMs on both Levy/Holt and AbstractATOMIC datasets in \cref{app:result_transferring}.

\begin{table}[t]
    \small
    \setlength{\tabcolsep}{3.1pt}
	\centering
	\begin{tabular}{l|ccccc}
	    \toprule
		\textbf{Models}&\textbf{B-1}&\textbf{B-2}&\textbf{R-2}&\textbf{R-L}&\textbf{Meteor}\\
            \midrule
            GPT2 &27.42&10.56&4.34&25.03&21.72 \\
            GPT2-medium &33.86&15.52&6.64&31.37&25.30 \\
            GPT2-large &49.23&29.64&16.80&48.36&35.44 \\
            GPT2-XL &53.90&32.39&\textbf{18.54}&53.73&\underline{38.45} \\
		\midrule
            GPT-J (6B) &55.65&31.19&15.20&54.42&36.70 \\
		Falcon (7B) &54.63&30.64&14.46&54.15&36.36 \\
            Falcon-Ins (7B) &53.18&30.15&14.96&51.90&35.17 \\
		  Llama2 (7B) &56.56&33.03&16.48&56.37&37.67 \\
		Llama2-Chat (7B) &57.11&34.42&16.31&54.87&37.34 \\
            Llama2 (13B) &\textbf{58.73}&\textbf{36.28}&\underline{17.63}&\textbf{57.45}&\textbf{39.47} \\
		Llama2-Chat (13B) &\underline{58.46}&\underline{34.54}&16.39&\underline{56.47}&37.95 \\
		\bottomrule
	\end{tabular}
	\caption{Results on the test set of \gendataname. B-1/2, R-2/L denote BLEU-1/2, ROUGE-2/L.}
        \label{tab:generation_all}
\end{table}

\section{Abstraction Generation Experiment}
In this section, we evaluate representative LMs on the \gendataname.

\subsection{Experiment Setup}
\paragraph{Evaluation Metric} BLEU-1, BLEU-2~\cite{DBLP:conf/acl/PapineniRWZ02}, ROUGE-2, ROUGE-L~\cite{lin-2004-rouge}, and Meteor~\cite{banerjee-lavie-2005-meteor} are computed to automatically evaluate all models.

\paragraph{Language Models} 
We evaluated representative LMs, including GPT-J (6B)~\cite{gpt-j}, Falcon (7B) and Falcon-Instruct (7B)~\cite{penedo2023refinedweb}, Llama2 (7B, 13B) and Llama2-Chat (7B, 13B)~\cite{touvron2023llama}, GPT2, and GPT2-medium/large/XL~\cite{radford2019language}. See implementation details in \cref{app:implementation_details}.

\subsection{Main Evaluation}

We present the overall performance of all language models in \cref{tab:generation_all}. We ascertain that fine-tuned language models can perform fairly well on our generation dataset. For example, Llama2 (13B) achieves the best BLEU-2 score, where 36.28\% of generated bi-grams are covered by the references. Unlike abstraction detection, increasing the number of parameters exerts a more significant effect on abstraction generation. For example, GPT2-XL (1.56B) gets the highest ROUGE-2 score, which is times higher than GPT2 (117M) and GPT2-medium (345M). Also, the performance of Llama2 (13B) is 1-3 points higher on all metrics than Llama2 (7B). Another noteworthy point is that instruction tuning does not help abstraction generation, exemplified by Llama2 (13B) getting higher metrics scores than Llama2-Chat (13B). We also include the performance on data of each entailment relation and conduct a human evaluation in \cref{app:generation_by_relation}. Similar to abstraction detection, we can find that models perform better on \nounrelation than other relations. Meanwhile, the human evaluation shows that automatic metrics highly correlate with human judgment. Then, we also list three kinds of generation errors of the fine-tuned Llama2 (13B) in \cref{app:generation_by_relation}.

\section{Conclusion}
In this paper, we introduce \papertitle to evaluate LLMs' abstraction ability. A scalable pipeline is designed to curate abstraction knowledge for three components of events. We carry out extensive experiments to demonstrate the comprehensiveness of our benchmark and provide valuable insights into the abstraction abilities of LLMs.

\section*{Limitations}
Our \papertitle incorporates extensive abstraction knowledge of events from ASER for nouns, verbs, and events. An open question is how to interleave the abstraction knowledge into the eventuality knowledge represented as explicit discourse relations in ASER. For the same event, we can have different levels of abstraction depending on the current context provided by eventuality knowledge. In the event ``I drink milk,'' ``milk'' can be abstracted as ``beverage'' under the situation that ``I am thirsty.'' In contrast, ``milk'' is better to be considered a kind of ``dairy product'' if ``I want to get more nutrition.'' Other knowledge can also be considered, such as factual knowledge~\cite{sun2023head} and commonsense knowledge~\cite{sap2019atomic, hwang2021comet, west2022symbolic}.

Representative LLMs are evaluated in our experiments. We leave for future work about building models with stronger abstraction abilities, including some sophisticated prompting methods~\cite{yao2023tree,long2023large,besta2023graph}, combining LLMs with smaller LMs~\cite{xu2023small}, semi-supervised learning~\cite{wang-etal-2023-cat}, retrieval augmented generation~\cite{lewis2020retrieval}.

\section*{Ethics Statement}
When constructing \papertitle, we sample head events from ASER~\cite{zhang2020aser,zhang2022aser}, an open-sourced eventuality graph. We only sampled eventualities extracted from Wikipedia and NYT, which are open-access. We carried out human annotation on Amazon Mechanical Turk (MTurk). Our payment rate is 1.2 USD for each HIT, which fulfills the minimum wage requirement and shows that annotators are fairly paid.

\section*{Acknowledgements}
The authors of this paper were supported by the NSFC Fund (U20B2053) from the NSFC of China, the RIF (R6020-19 and R6021-20) and the GRF (16211520 and 16205322) from RGC of Hong Kong. We also thank the support from the Tencent AI Lab Rhino-Bird Focused Research Program and the UGC Research Matching Grants (RMGS20EG01-D, RMGS20CR11, RMGS20CR12, RMGS20EG19, RMGS20EG21, RMGS23CR05, RMGS23EG08).

\newpage
\bibliography{anthology,custom}
\bibliographystyle{acl_natbib}

\newpage
\appendix
\section{Data Curation Details}
\subsection{ASER Cleaning}
\label{app:aser_cleaning}
Since ASER is an eventuality graph automatically extracted from diverse corpora, some noisy extraction results exist. Thus, we design a few rules to clean some frequent noise categories in ASER.

First, we found that many eventualities are noisy due to incompleteness. For example, ``the norman army weakened,'' an eventuality extracted from Wikipedia, misses the linking verb ``was'' in the passive voice. To solve this, we re-parse each eventuality and remove eventualities whose dependency graph changes in the re-parsing stage. With this rule, we remove a lot of incomplete eventualities. 

Then, we design four lexical rules for noisy eventualities: (1) We find that many eventualities with the \texttt{s-v} pattern (see \cite{zhang2022aser} for definition) contain light verbs. We remove those eventualities since they lack semantic meanings, such as ``they do.'' (2) We find that the parsing algorithm of ASER can extract eventualities from subordinate clauses but cannot link relatives to antecedents. For example, ``who won the competition'' is extracted from the sentence ``Bob is a painter who won the competition'' without replacing ``who'' with ``Bob.'' We remove all eventualities starting with relatives. (3) ASER also contains some eventualities that are totally composed of stopwords. We remove them since they also do not have too many semantic meanings, such as ``She just won.'' (4) We remove eventualities containing URLs and HTML tags.

In detail, the light verbs we use are \texttt{do}, \texttt{give}, \texttt{have}, \texttt{make}, \texttt{get}, and \texttt{take}, as well as their inflections, such as \texttt{doing} and \texttt{has}. The relatives we use are \texttt{how}, \texttt{what}, \texttt{when}, \texttt{where}, \texttt{which}, \texttt{who}, \texttt{why}, \texttt{whatever}, \texttt{whose}, \texttt{whom}, and \texttt{if}. The stopword list is accessed by NLTK~\cite{bird2009nltk}.

\subsection{Matching Nouns and Verbs}
\label{app:matching_nouns_and_verbs}
In our benchmark, the abstraction knowledge of \nounrelation and \verbrelation involves identifying nouns and verbs from events. In ASER, each word in the syntactic pattern is classified into word types according to their POS tags, including \texttt{noun}, \texttt{verb}, \texttt{be}, and \texttt{preposition}. We use those word types to identify the nouns and verbs. For example, the pattern \texttt{subject-verb-object} has word types \texttt{noun}, \texttt{verb}, and \texttt{noun} for each word. Also, we identify modifiers to complete each noun by collecting all words dependent on the noun in the dependency parsing graph, such as ``fluffy'' in ``fluffy cat.''

We also take care of some special cases where eventualities contain some transparent nouns~\cite{meyersannotation}, such as ``I have \textbf{a lot of food}.'' In this case, we identify ``food'' as an instance instead of ``lot.'' Verbs also have similar constructions, such as ``I \textbf{am going to sleep}.'' In this example, we identify ``sleep'' as an instance instead of ``going.''

\subsection{Prompts for Collecting Data}
\label{app:prompts_abstract_concept}
We provide the prompt template used in collecting abstract concepts in \cref{app:abstract_concept_prompt} and the prompt template used in collecting negative examples in \cref{app:negative_example_prompt}.

\begin{table}[t!]
\centering
\small

\begin{subtable}[t]{0.96\columnwidth}
\centering
\begin{tabular}{p{0.96\columnwidth}}
	\toprule
	\textbf{Task Instruction:} In this task, you need to list the hypernyms of an instance. Hypernyms are words that represent broader categories or concepts. \\
	\midrule
	\textbf{Exemplar Input:} 1. Given the sentence ``the clinic had resumed its work,'' what is the list of hypernyms of ``clinic?''\\
	\textbf{Exemplar Output:} (1) medical facility, (2) healthcare center, \ldots, (10) diagnostic center. \\
    \midrule
    \textbf{Following Exemplars:} Exemplar 2, Exemplar 3, \ldots, Exemplar 10 \\
    \midrule
    \textbf{Testing Input:} 11. Given the sentence \textbf{[HEAD]}, what is the list of hypernyms of \textbf{[INSTANCE]}? \\
    \bottomrule
	\end{tabular}
\caption{\nounrelation}
\end{subtable}

\begin{subtable}[t]{0.96\columnwidth}
\centering
\begin{tabular}{p{0.96\columnwidth}}
	\toprule
	\textbf{Task Instruction:} In this task, you need to list some abstract descriptions of an event. \\
	\midrule
	\textbf{Exemplar Input:} 1. Which abstract descriptions can the event ``PersonX surfs the web'' be summarized as?\\
	\textbf{Exemplar Output:} (1) surfing, (2) surfing the internet, \ldots, (10) browsing the internet. \\
    \midrule
    \textbf{Following Exemplars:} Exemplar 2, Exemplar 3, ..., Exemplar 10 \\
    \midrule
    \textbf{Testing Input:} 11. Which abstract descriptions can the event \textbf{[HEAD]} be summarized as? \\
    \bottomrule
	\end{tabular}
\caption{\eventrelation}
\end{subtable}
\caption{The prompt we used to collect abstract concepts from ChatGPT for \nounrelation and \eventrelation relations. Two placeholders \textbf{[HEAD]} and \textbf{[ISNTANCE]} will be replaced with real head events and instances. We present the prompt in the dialogue format. Please concatenate all utterances to form the prompt of GPT3.5.}
\label{app:abstract_concept_prompt}
\end{table}

\begin{table}[t!]
\centering
\small
\begin{subtable}[t]{0.96\columnwidth}
\centering
\begin{tabular}{p{0.96\columnwidth}}
	\toprule
	\textbf{Task Instruction:} In this task, you need to list some related nouns but not hypernyms. Hypernyms are words that represent broader categories or concepts. \\
	\midrule
	\textbf{Exemplar Input:}  1. Given the sentence ``the clinic had resumed its work,'' please list related nouns of ``clinic'' but not hypernyms.\\
	\textbf{Exemplar Output:} (1) patients, (2) doctors, \ldots, (10) mask. \\
    \midrule
    \textbf{Following Exemplars:} Exemplar 2, Exemplar 3, ..., Exmplar 10 \\
    \midrule
    \textbf{Testing Input:} 
    11. Given the sentence \textbf{[HEAD]}, please list related nouns of \textbf{[INSTANCE]} but not hypernyms. \\
    \bottomrule
	\end{tabular}
\caption{\nounrelation}
\end{subtable}

\begin{subtable}[t]{0.96\columnwidth}
\centering
\begin{tabular}{p{0.96\columnwidth}}
	\toprule
	\textbf{Task Instruction:} In this task, you need to list some related phrases but not abstract descriptions of an event. \\
	\midrule
	\textbf{Exemplar Input:} 1. Please list related phrases of the event ``PersonX surfs the web'' but not abstract descriptions of it. \\
	\textbf{Exemplar Output:} (1) typing a URL, (2) website, \ldots, (10) bandwidth. \\
    \midrule
    \textbf{Following Exemplars:} Exemplar 2, Exemplar 3, ..., Exemplar 10 \\
    \midrule
    \textbf{Testing Input:}
    11. Please list related phrases of the event \textbf{[HEAD]} but not abstract descriptions of it. \\
    \bottomrule
	\end{tabular}
\caption{\eventrelation}
\end{subtable}
\caption{The prompt we used to collect challenging negative examples from ChatGPT for \nounrelation and \eventrelation relations.}
\label{app:negative_example_prompt}
\end{table}

\subsection{Annotation Details}
\label{app:annotation_details}
There are two qualification tests to choose workers to maintain rigorous quality control. First, we invited annotators who meet the following conditions to take our qualification examinations: 1) an approval rate of above 95\% and 2) at least a thousand approved HITs. In the second round, qualification questions, including effortless and tricky examples, are collected by this paper's authors, who clearly understand abstract tuples. The experts annotate 200 tuples for each relation. An annotator should correctly answer 18 of 20 questions to pass the second round test. 

\begin{figure*}[h]
    \centering
    \includegraphics[width=2\columnwidth]{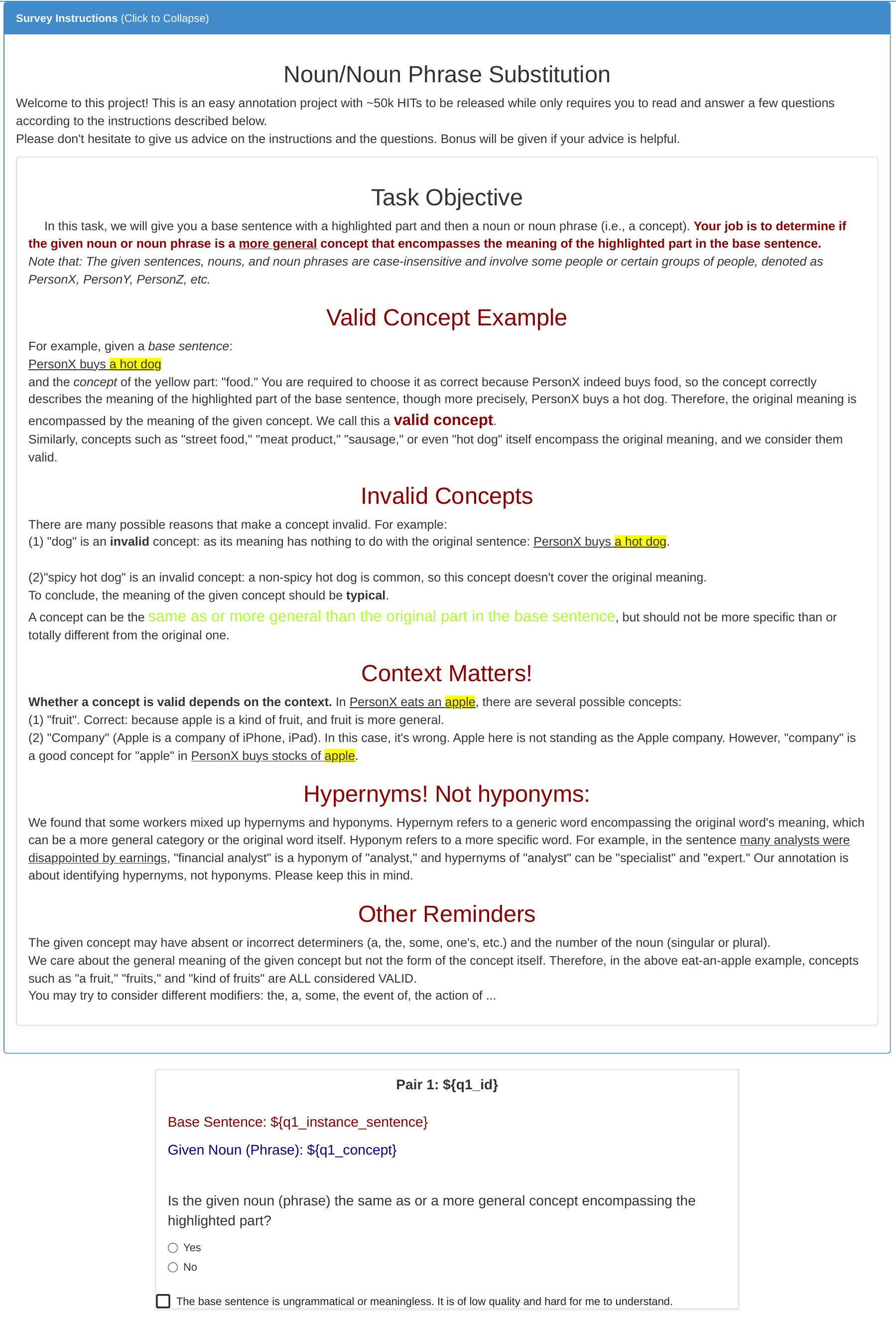}
    \caption{The full text of instructions provided to annotators on Amazon Mechanical Turk (MTurk). There are ten questions in a Human Intelligence Task (HIT), and we only display one here for brevity.}
    \label{fig:full_instruction}
\end{figure*}

In our main annotation, we assign each tuple to 5 annotators in the first round of annotations. We manually inspect their annotation quality and disqualify those annotators who cannot continue to annotate with high accuracy. The annotations from those disqualified annotators are then discarded for quality control. For higher quality, we also introduce two rounds of refinement. We reannotate the discarded votes in the first round of refinement. In the second round, we request annotators to reannotate the tuples that do not reach an agreement (i.e., 2 or 3 out of 5 annotators vote for valid). After this, we discard examples that annotators still do not agree on. We show the full text of instructions provided to annotators in \cref{fig:full_instruction}.

During our massive annotation process, 5153 annotators participated in qualification tests, with 551 (10.7\%) annotators passing them. The IAA score of pairwise agreement proportion is 77.62\%, and Fleiss’s $\kappa$~\cite{fleiss1971measuring} is 0.54.

\begin{table}[t]
    \small
    \setlength{\tabcolsep}{3pt}
	\centering
	\begin{tabular}{lcccccc}
	\toprule
        \multirow{2}{*}{\textbf{LLMs}}&\multicolumn{2}{c}{\textbf{Noun}} &\multicolumn{2}{c}{\textbf{Verb}}&\multicolumn{2}{c}{\textbf{Event}}\\
        \cmidrule(rl){2-3}\cmidrule(rl){4-5}\cmidrule(rl){6-7}
	&\textbf{Acc}&\textbf{Ma-F1} &\textbf{Acc}&\textbf{Ma-F1} & \textbf{Acc}&\textbf{Ma-F1} \\
            \midrule
            GPT 4&62.70&62.47&\textcolor{red}{57.70}&\textcolor{red}{57.54}&66.20&64.06\\
            GPT 3.5&66.10&\textcolor{red}{62.72}&54.10&53.94&\textcolor{red}{67.40}&\textcolor{red}{59.57}\\
            \midrule
            ChatGPT&67.40&66.04&55.20&55.04&67.60&\textcolor{red}{63.36}\\
            \multicolumn{1}{r}{+ CoT} &56.70&56.67&54.00&52.39&61.30&60.13\\
		\bottomrule
	\end{tabular}
	\caption{Results of \emph{NLI} prompt on \detectiondataname. We mark scores higher than scores of \emph{Abs.} prompt in \cref{tab:main_eval_test} with red color. We can see that most scores are inferior.}
    \label{tab:prompt_robustness}
\end{table}

\begin{table}[t!]
\centering
\small
\begin{subtable}[t]{0.96\columnwidth}
\centering
\begin{tabular}{p{0.96\columnwidth}}
	\toprule
    \textbf{\nounrelation, \verbrelation and \eventrelation: } Identify entailment and provide a ``Yes'' or ``No'' response. Entailment is about determining whether a ``hypothesis'' is true given a ``premise.'' Given the premise \textbf{[HEAD]}, can we know the hypothesis \textbf{[TAIL]}? \\
    \bottomrule
	\end{tabular}
\caption{Zero-Shot Prompt}
\end{subtable}

\begin{subtable}[t]{0.96\columnwidth}
\centering
\begin{tabular}{p{0.96\columnwidth}}
	\toprule
	\textbf{\nounrelation, \verbrelation and \eventrelation:} Identify entailment, which is about determining whether a ``hypothesis'' is true given a ``premise.'' Given the premise \textbf{[HEAD]}, can we know the hypothesis \textbf{[TAIL]}? Step 1: Let's think about meanings of those sentences. Step 2: Provide a ``Yes'' or ``No'' response. \\
    \bottomrule
	\end{tabular}
\caption{CoT Prompt}
\end{subtable}
\caption{The NLI-format prompt. Results of this prompt is shown in \cref{tab:prompt_robustness}. Placeholders \textbf{[HEAD]} and \textbf{[TAIL]} will be replaced with real head events and tail events.}
\label{tab:detection_NLI_prompt}
\end{table}


\begin{table}[t!]
\centering
\small
\begin{subtable}[t]{0.96\columnwidth}
\centering
\begin{tabular}{p{0.96\columnwidth}}
	\toprule
	\textbf{\nounrelation:} Identify the hypernym of a specific noun and provide a ``Yes'' or ``No'' response. Hypernyms are words with a broad meaning, which more specific words fall under. In the sentence \textbf{[HEAD]}, does the meaning of \textbf{[CONCEPT]} encompass \textbf{[INSTANCE]}? \\
    \midrule
    \textbf{\verbrelation: } Identify the hypernym of a specific verb and provide a ``Yes'' or ``No'' response.
    Hypernyms are words with a broad meaning, which more specific words fall under. In the sentence \textbf{[HEAD]}, does the meaning of \textbf{[CONCEPT]} encompass \textbf{[INSTANCE]}? \\
    \midrule
    \textbf{\eventrelation: } Identify abstract descriptions of specific sentences, and provide a ``Yes'' or ``No'' response. Can we consider \textbf{[CONCEPT]} as an abstract description of the sentence \textbf{[HEAD]}? \\
    \bottomrule
	\end{tabular}
\caption{Zero-Shot Prompt}
\end{subtable}

\begin{subtable}[t]{0.96\columnwidth}
\centering
\begin{tabular}{p{0.96\columnwidth}}
	\toprule
    \textbf{\nounrelation:} Identify the hypernym of a specific noun. Hypernyms are words with a broad meaning, which more specific words fall under. In the sentence \textbf{[HEAD]}, does the meaning of \textbf{[CONCEPT]} encompass \textbf{[INSTANCE]}? Step 1: Let's think about the meanings of those words. Step 2: Provide a ``Yes'' or ``No'' response. \\
    \midrule
    \textbf{\verbrelation: } Identify the hypernym of a specific verb. Hypernyms are words with a broad meaning, which more specific words fall under. In the sentence \textbf{[HEAD]}, does the meaning of \textbf{[CONCEPT]} encompass \textbf{[INSTANCE]}? Step 1: Let's think about the meanings of those words. Step 2: Provide a ``Yes'' or ``No'' response. \\
    \midrule
    \textbf{\eventrelation: } Identify abstract descriptions of specific sentences. Can we consider \textbf{[CONCEPT]} as an abstract description of the sentence \textbf{[HEAD]}? Step 1: Let's think about the meanings of the sentence and the abstract description. Step 2: Provide a ``Yes'' or ``No'' response. \\
    \bottomrule
	\end{tabular}
\caption{CoT Prompt}
\end{subtable}
\caption{The default prompt we used (i.e., \emph{Abs.} prompt) to test GPT3.5, ChatGPT, and GPT4. The results of this prompt are shown in \cref{tab:main_eval_test}. Placeholders \textbf{[HEAD]}, \textbf{[INSTANCE]}, and \textbf{[CONCEPT]} will be replaced with real head events, instances, and abstract concepts.}
\label{tab:detection_abs_prompt}
\end{table}

\section{Implementation Details}
\label{app:implementation_details}
First, we discuss details shared in both abstraction detection and abstraction generation experiments.
We access open-source language models using Transformers~\cite{wolf2020transformers} and fine-tune them on 8 NVIDIA A100 (80G) GPUs. LLMs with 7B and 13B parameters are loaded with BF16. The best checkpoint is selected according to the sum of all metrics on the validation set. When fine-tuning LLMs with LoRA, we only add new parameters to attention layers with the rank and $\alpha$ equal to 64 and 128. We grid search the learning rate of 5e-6, 1e-5, 5e-5, and batch sizes of 64 and 128.

Here are some details specific to abstraction detection experiments.
When fine-tuning \textbf{NLI models}, we re-use the classification layer with ``Entailment'' and ``Neutral'' for valid and invalid, respectively. We access ChatGPT, GPT4, and GPT3.5 via OpenAI API\footnote{https://platform.openai.com/docs/api-reference}, with specific versions being \texttt{gpt-3.5-turbo-0613}, \texttt{gpt-4-0613}, and \texttt{gpt-3.5-turbo-instruct-0914}. They are evaluated on one thousand examples that we randomly sampled from the testing set of each relation due to the trade-off between API expenses and our evaluation's precision. In addition, we provide ChatGPT with ten exemplars for in-context learning.

\section{Experimental Results}
In this appendix, we collect supplementary abstraction detection and generation results.

\subsection{Validation Results on Abstraction Detection}
\label{app:validation_results}
We collect the performance of LMs trained on each entailment relation separately on the validation set of the \detectiondataname in \cref{tab:main_eval_validation}. Then, we present the performance of LMs trained on merged data of all entailment relations on the validation set in \cref{tab:multi_relation_validation}.

\subsection{ChatGPT Prompt Robustness}
\label{app:chatgpt_prompt_robustness}
First, we ask GPT3.5, ChatGPT, and GPT4 whether an abstract concept is valid as the default prompt (denoted as \emph{Abs.} prompt). The prompt is presented in \cref{tab:detection_abs_prompt}, and its results are shown in \cref{tab:main_eval_test}. Meanwhile, we design another prompt in NLI format, treating the head and tail events as the premise and hypothesis (denoted as \emph{NLI} prompt). This prompt is presented in \cref{tab:detection_NLI_prompt}. As shown in \cref{tab:prompt_robustness}, the performance of the \emph{NLI} prompt is inferior to the \emph{Abs.} prompt on most metrics, showing the robustness of the \emph{Abs.} prompt.

\begin{table}[t]
    \small
	\centering
	\begin{tabular}{l||cccc}
	    \toprule
		\textbf{Models}&\textbf{Acc}&\textbf{Ma-F1}&\textbf{AUC}&\textbf{APS}\\
		\midrule
            Falcon (7B) &82.93&74.57&86.55&57.46 \\
            Mistral (7B) &84.56&76.67&88.60&62.78 \\
            Llama2 (7B) &84.20&74.81&87.75&60.98 \\
            Llama2 (13B) &84.47&76.28&86.27&58.69 \\
            \midrule
            CF-Falcon (7B) &87.19&80.52&91.21&71.21\\
            CF-Mistral (7B) &88.28&82.14&92.64&77.78\\
            CF-Llama2 (7B) &88.55&83.04&92.83&75.83\\
            CF-Llama2 (13B) &87.70&81.48&92.33&74.51\\
		\bottomrule
	\end{tabular}
	\caption{The fine-tuning performance of LLMs on the Levy/Holt dataset. \textbf{CF} stands for continually fine-tuning.}
	\label{tab:full_finetune_levy_holt}
\end{table}

\begin{table}[t]
    \small
	\centering
	\begin{tabular}{l||cc}
	    \toprule
		\textbf{Models}&\textbf{Acc}&\textbf{Ma-F1}\\
		\midrule
            Falcon-Ins (7B) & 73.30 & 42.66 \\
            Mistral-Ins (7B) & 72.40 & 57.81 \\
            Llama2-Chat (7B) & 71.30 & 45.65 \\
            Llama2-Chat (13B) & 71.70 & 42.77 \\
		\bottomrule
	\end{tabular}
	\caption{The zero-shot performance of instruction-tuned LLMs on the Levy/Holt dataset.}
	\label{tab:instruction_tuned_levy_holt}
\end{table}

\begin{table}[t]
    \small
	\centering
	\begin{tabular}{p{1.7cm}|l|ccc}
	    \toprule
		\textbf{Models} & \textbf{Shot}&\textbf{Acc}&\textbf{Ma-F1}&\textbf{AUC}\\
            \midrule
		\multirow{5}{*}{Falcon (7B)}
            & 0\% &59.39&41.01&61.18 \\
            & 20\% &73.41&72.36&80.20 \\
            & 40\% &81.17&80.36&88.73 \\
		& 60\% &82.37&81.76&89.73 \\
	    & 80\% &83.13&82.71&91.20 \\
            \midrule
		\multirow{5}{*}{Mistral (7B)}
            & 0\% &41.88&31.44&53.71 \\
            & 20\% &83.14&82.64&90.56 \\
            & 40\% &84.12&83.90&92.57 \\
		& 60\% &85.66&85.30&92.98 \\
	    & 80\% &85.72&85.42&93.66 \\
	    \midrule
		\multirow{5}{*}{Llama2 (7B)}
            & 0\%  &59.39&41.01&61.18 \\
            & 20\% &80.28&79.61&87.89 \\
            & 40\% &82.93&82.33&90.96 \\
		& 60\% &83.12&82.76&91.41 \\
	    & 80\% &85.67&85.19&92.97 \\
            \midrule
		\multirow{5}{*}{Llama2 (13B)}
            & 0\% &55.94&38.81&43.41 \\
            & 20\% &75.59&74.56&82.19 \\
            & 40\% &81.87&81.30&89.71 \\
		& 60\% &82.98&82.28&90.44 \\
	    & 80\% &84.93&84.31&92.39 \\
		\bottomrule 
	\end{tabular}
	\caption{The few-shot performance on the test set of AbstractATOMIC dataset. LLMs are loaded from pre-trained configurations.}
	\label{tab:abstract_atomic_pretrain}
\end{table}

\subsection{Full Results of Transferring to Other Sources}
\label{app:result_transferring}
For the zero-shot study on the Levy/Holt dataset, we also provide the zero-shot performance of instruction-tuned LLMs for a complete comparison. As shown in \cref{tab:instruction_tuned_levy_holt}, the performance of instruction-tuned models is much lower than models fine-tuned on our benchmark, showing the comprehensiveness of our benchmark. 

Meanwhile, the full fine-tuning performance of all LLMs on the Levy/Holt dataset is shown in \cref{tab:full_finetune_levy_holt}. Also, we provide the full results of all pre-trained LLMs on AbstractATOMIC in \cref{tab:abstract_atomic_pretrain} and results of LLMs that initially fine-tuned on our dataset in \cref{tab:abstract_atomic_continual_finetune}.

\begin{table}[H]
    \small
	\centering
	\begin{tabular}{p{1.7cm}|l|ccc}
	    \toprule
		\textbf{Models} & \textbf{Shot}&\textbf{Acc}&\textbf{Ma-F1}&\textbf{AUC}\\
		\midrule
            \multirow{5}{*}{Falcon (7B)}
            & 0\% &64.22&64.22&72.80 \\
            & 20\% &81.11&80.54&89.01 \\
            & 40\% &83.49&82.98&91.11 \\
		& 60\% &83.95&83.45&91.66 \\
	    & 80\% &84.67&84.22&92.24 \\
            \midrule
		\multirow{5}{*}{Mistral (7B)} & 0\% &64.81&64.78&73.60 \\
            & 20\% &84.43&84.03&91.73 \\
            & 40\% &85.85&85.40&92.88 \\
		& 60\% &86.24&85.75&93.23 \\
	    & 80\% &86.61&86.20&93.71 \\
	    \midrule
		\multirow{5}{*}{Llama2 (7B)} & 0\% &62.40&62.13&71.65 \\
            & 20\% &82.70&82.32&90.43 \\
            & 40\% &84.51&84.06&91.90 \\
		& 60\% &84.91&84.50&92.26 \\
	    & 80\% &85.97&85.59&93.13 \\
            \midrule
		\multirow{5}{*}{Llama2 (13B)} & 0\% &64.28&64.25&71.35 \\
            & 20\% &82.76&82.30&90.23 \\
            & 40\% &84.50&84.00&91.88 \\
		& 60\% &84.91&84.48&92.22 \\
	    & 80\% &85.87&85.46&93.01 \\
		\bottomrule 
	\end{tabular}
	\caption{The few-shot performance on the test set of AbstractATOMIC dataset. LLMs are initially trained on \detectiondataname.}
	\label{tab:abstract_atomic_continual_finetune}
\end{table}

\subsection{Full Results of Abstraction Generation}
\label{app:generation_by_relation}
To carry out a more thorough evaluation of LMs' ability to generate abstraction knowledge, we also provide performance by entailment relations \nounrelation, \verbrelation, and \eventrelation in \cref{tab:generation_noun,tab:generation_verb,tab:generation_event}, respectively. 

Meanwhile, we conduct the human evaluation of GPT2 and Llama2 (13B) on 50 examples for each relation (150 in total). The annotation is conducted by an expert about whether a given generated concept is valid. From the results in \cref{tab:human_eval}, we can find that the automatic evaluation results correlate with the human evaluation, showing the effectiveness of the automatic metrics.

Further, we also provide error analyses of three concepts generated by Llama2 (13B), shown in \cref{tab:more_error_analysis}. These cases show that fine-tuned LLMs can be wrong when (1) generating word meanings instead of concepts, (2) repeating the given instance, and (3) generating related phrases (but not abstract concepts). 

\begin{table}[t!]
\centering
\small
\begin{tabular}{p{0.96\columnwidth}}
\toprule
\multicolumn{1}{c}{\textbf{Example \#1}} \\ \midrule
\textbf{Head Event}: PersonX snared the important wicket of PersonY. \\
\textbf{Instance}: important wicket of PersonY \\ \textbf{Entailment Relation}: \nounrelation \\
\textbf{Generated Concept}: This means the wicket of PersonY \\
\textbf{Expert Explanation}: The generation is an explanation of the meaning instead of some abstract concepts. \\
\midrule \midrule
\multicolumn{1}{c}{\textbf{Example \#2}} \\ \midrule
\textbf{Head Event}: PersonX lived for decades. \\
\textbf{Instance}: lived \\ 
\textbf{Entailment Relation}: \verbrelation \\
\textbf{Generated Concept}: lived \\
\textbf{Expert Explanation}: The generation is the instance itself, not an abstract concept for it. \\
\midrule \midrule
\multicolumn{1}{c}{\textbf{Example \#3}} \\ \midrule
\textbf{Head Event}: Each squadron meets its specific mission-oriented needs. \\
\textbf{Instance}: each squadron meets its specific mission-oriented needs \\ 
\textbf{Entailment Relation}: \eventrelation \\
\textbf{Generated Concept}: mission-specific requirements \\
\textbf{Expert Explanation}: The sentence emphasizes that the needs are met, not only the needs themselves. So, a correct generation should be "requirement satisfaction," "needs fulfillment," etc.
\\
\bottomrule
\end{tabular}

\caption{Error analysis of generated concepts from Llama2 (13B).}
\label{tab:more_error_analysis}
\end{table}

\begin{table}[t]
    \small
	\centering
	\begin{tabular}{l|cccc}
	    \toprule
		\textbf{Models}&\textbf{Noun}&\textbf{Verb}&\textbf{Event}&\textbf{All}\\
            \midrule
            GPT2 & 48.00 & 26.00 &	44.00 & 39.33 \\
            Llama2 (13B) & 90.00 &	66.00 & 74.00 & 76.67 \\
		\bottomrule
	\end{tabular}
	\caption{Human evaluation of GPT2 and Llama2 (13B).}
        \label{tab:human_eval}
\end{table}

\begin{table}[t]
    \small
    \setlength{\tabcolsep}{3.1pt}
	\centering
	\begin{tabular}{l|ccccc}
	    \toprule
		\textbf{Models}&\textbf{B-1}&\textbf{B-2}&\textbf{R-2}&\textbf{R-L}&\textbf{Meteor}\\
            \midrule
            GPT2 & 33.67&11.63&3.35&30.75&20.04\\
            GPT2-medium &39.15&15.64&6.09&39.43&24.82 \\
            GPT2-large &55.79&30.16&15.18&57.31&37.93 \\
            GPT2-XL &62.47&33.94&18.70&64.67&42.30 \\
		\midrule
            GPT-J (6B) &67.47&35.65&15.47&67.17&41.32 \\
		Falcon (7B) &68.67&36.48&16.25&71.62&43.63 \\
            Falcon-Ins (7B) &63.92&32.08&13.51&65.31&39.49 \\
		  Llama2 (7B) &65.80&33.73&17.28&70.29&43.47 \\
		Llama2-Chat (7B) &70.07&39.08&18.12&71.51&45.00 \\
            Llama2 (13B) &68.81&34.91&18.02&71.04&45.17 \\
		Llama2-Chat (13B) &68.71&33.60&16.67&70.54&43.79 \\
		\bottomrule
	\end{tabular}
	\caption{Generation results on data of \nounrelation in the test set of \gendataname. B-1/2, R-2/L denote BLEU-1/2, ROUGE-2/L, respectively.}
    \label{tab:generation_noun}
\end{table}

\begin{table}[t]
    \small
    \setlength{\tabcolsep}{3.1pt}
	\centering
	\begin{tabular}{l|ccccc}
	    \toprule
		\textbf{Models}&\textbf{B-1}&\textbf{B-2}&\textbf{R-2}&\textbf{R-L}&\textbf{Meteor}\\
            \midrule
            GPT2 &5.44&0.00&0.00&5.79&18.21 \\
            GPT2-medium &11.46&1.25&0.18&11.77&21.00 \\
            GPT2-large &40.34&44.37&12.23&36.98&30.58 \\
            GPT2-XL &44.14&39.47&10.77&42.62&31.99 \\
		\midrule
            GPT-J (6B) &40.82&31.46&5.11&40.33&27.66 \\
		Falcon (7B) &36.88&28.77&3.83&37.01&26.06 \\
            Falcon-Ins (7B) &38.49&38.38&6.93&36.68&26.30 \\
		  Llama2 (7B) &43.92&36.47&5.29&41.94&27.45 \\
		Llama2-Chat (7B) &36.68&26.58&3.83&36.79&24.32 \\
            Llama2 (13B) &45.18&43.53&6.75&43.90&29.85 \\
		Llama2-Chat (13B) &42.25&35.16&5.84&41.94&27.76 \\
		\bottomrule
	\end{tabular}
	\caption{Generation results on data of \verbrelation in the test set of \gendataname. B-1/2, R-2/L denote BLEU-1/2, ROUGE-2/L, respectively.}
 \label{tab:generation_verb}
\end{table}

\begin{table}[t]
    \small
    \setlength{\tabcolsep}{3.1pt}
	\centering
	\begin{tabular}{l|ccccc}
	    \toprule
		\textbf{Models}&\textbf{B-1}&\textbf{B-2}&\textbf{R-2}&\textbf{R-L}&\textbf{Meteor}\\
            \midrule
            GPT2 &35.24&10.93&10.86&42.19&28.06 \\
            GPT2-medium &44.12&17.54&15.28&46.23&31.19 \\
            GPT2-large &50.39&25.52&24.38&52.01&38.57 \\
            GPT2-XL &53.92&29.73&27.98&54.69&41.96 \\
		\midrule
            GPT-J (6B) &56.28&29.24&27.38&56.96&42.51 \\
		Falcon (7B) &55.15&28.24&25.53&54.96&40.63 \\
            Falcon-Ins (7B) &54.90&27.88&26.63&55.10&41.10 \\
		  Llama2 (7B) &57.48&32.16&29.40&58.00&43.56 \\
		Llama2-Chat (7B) &60.18&33.52&29.66&57.84&44.51 \\
            Llama2 (13B) &59.34&35.82&30.66&58.36&44.74 \\
		Llama2-Chat (13B) &61.06&34.88&29.13&58.04&43.74 \\
		\bottomrule
	\end{tabular}
	\caption{Generation results on data of \eventrelation in the test set of \gendataname. B-1/2, R-2/L denote BLEU-1/2, ROUGE-2/L, respectively.}
 \label{tab:generation_event}
\end{table}

\begin{table*}[t!]
\setlength{\tabcolsep}{4pt}
\centering
\small
\begin{tabular}{l||ccc|ccc|ccc|ccc}
	\toprule
        \multirow{2}{*}{\textbf{LLM + LoRA}}&\multicolumn{3}{c|}{\textbf{Noun}} &\multicolumn{3}{c|}{\textbf{Verb}}&\multicolumn{3}{c|}{\textbf{Event}}&\multicolumn{3}{c}{\textbf{All}}\\
	&\textbf{Acc}&\textbf{Ma-F1} &\textbf{AUC} &\textbf{Acc}&\textbf{Ma-F1} &\textbf{AUC}&\textbf{Acc}&\textbf{Ma-F1} &\textbf{AUC} &\textbf{Acc}&\textbf{Ma-F1} &\textbf{AUC}  \\
		\midrule
            Falcon (7B) &88.12&87.55&92.60&64.42&64.15&68.92&77.54&71.84&80.38&78.76&77.38&85.95\\
            Falcon-Ins (7B) &87.62&87.09&92.44&64.61&64.59&69.23&77.39&71.44&80.29&78.52&77.37&85.88\\
            Mistral (7B) &88.90&88.38&92.86&64.61&64.30&69.75&77.95&72.56&81.07&79.28&77.96&86.73\\
            Mistral-Ins (7B) &88.57&88.09&92.77&64.49&64.40&68.76&77.78&72.10&81.02&79.04&77.86&86.50\\
            Llama2 (7B)&88.85&88.29&92.97&64.17&63.84&68.95&77.97&71.95&80.97&79.15&77.71&86.59\\
            Llama2-Chat (7B)&88.37&87.82&92.86&64.07&63.94&68.93&77.39&71.53&79.68&78.78&77.82&86.04\\
            Llama2 (13B)&88.26&87.83&92.85&65.20&65.20&69.48&77.65&71.95&80.57&79.06&78.08&86.57\\
            Llama2-Chat (13B)&88.62&88.09&92.77&65.47&65.31&69.71&77.65&72.11&81.31&79.25&78.01&86.60\\
		\bottomrule
\end{tabular}
	\caption{The performance of LLMs on the validation set of \detectiondataname under the multi-relation setting.}
 \label{tab:multi_relation_validation}
\end{table*}

\begin{table*}[t]
    \small
    \setlength{\tabcolsep}{4.9pt}
	\centering
	\begin{tabular}{l|l||ccc|ccc|ccc}
	\toprule
        \multirow{2}{*}{\textbf{Methods}}&\multirow{2}{*}{\textbf{Backbone}}&\multicolumn{3}{c|}{\textbf{Noun}} &\multicolumn{3}{c|}{\textbf{Verb}}&\multicolumn{3}{c}{\textbf{Event}}\\
	&&\textbf{Acc}&\textbf{Ma-F1} &\textbf{AUC} &\textbf{Acc}&\textbf{Ma-F1} &\textbf{AUC} & \textbf{Acc}&\textbf{Ma-F1}&\textbf{AUC} \\
            \midrule
            \textbf{Random} & \multicolumn{1}{|c||}{-} & 50.00 & 49.67 & 50.00 & 50.00 & 49.97 & 50.00 & 50.00 & 49.01 & 50.00\\
            \textbf{Majority Vote} & \multicolumn{1}{|c||}{-} & 58.11 & - & 50.00 & 52.40 & - & 50.00 & 63.94 & - & 50.00 \\
		\midrule
            \multirow{4}{*}{\textbf{NLI + Zero}}&BART-large-mnli&70.44&67.65&75.47&54.84&45.89&62.54&71.32&66.65&71.06\\
            &RoBERTa-large-mnli&67.76&62.61&74.70&54.10&43.55&61.51&70.40&62.65&70.62\\
            &DeBERTa-base-mnli&67.77&65.05&72.35&54.72&46.35&61.34&66.14&62.52&67.21\\
            &DeBERTa-large-mnli&72.85&70.95&78.23&55.68&48.23&62.34&68.35&65.30&70.55\\
            \midrule
            \multirow{4}{*}{\textbf{NLI + FT}}&BART-large-mnli&86.47&86.03&91.92&64.47&64.47&68.53&75.58&71.02&79.63\\
            &RoBERTa-large-mnli&86.93&86.35&91.92&65.16&64.83&69.06&77.75&71.42&80.25\\
            &DeBERTa-base-mnli&86.17&85.42&91.24&64.64&64.61&68.96&77.36&70.66&79.50\\
            &DeBERTa-large-mnli&86.92&86.30&91.78&64.15&64.08&69.30&77.47&71.07&79.65\\
            \midrule
            \multirow{6}{*}{\textbf{PLM + FT}}&BERT-base&85.47&84.78&91.02&63.38&63.32&68.35&77.33&71.06&80.27\\
            &BERT-large&86.65&86.03&91.37&62.96&62.95&67.02&76.16&70.84&79.73\\
            &RoBERTa-base&85.01&84.31&90.76&62.62&62.61&67.04&77.25&71.37&79.75\\
            &RoBERTa-large&86.35&85.80&91.29&62.91&62.91&67.64&77.86&71.53&79.89\\
            &DeBERTa-base&85.22&84.51&90.31&62.28&61.89&67.34&76.85&71.25&79.55\\
            &DeBERTa-large&87.77&87.23&91.91&64.79&64.79&68.49&77.75&71.58&80.05\\
            \midrule
            \multirow{8}{*}{\textbf{LLM + LoRA}}            &Falcon (7B)&87.49&86.97&92.33&63.56&63.43&68.13&76.45&71.49&79.50\\
            &Falcon-Ins (7B)&86.57&86.11&92.07&64.15&64.09&68.46&76.17&70.53&78.89\\
            &Mistral (7B)&88.50&88.08&92.63&63.29&62.90&68.16&77.91&71.52&80.58\\
            &Mistral-Ins (7B)&88.31&87.90&92.60&63.71&63.65&68.77&77.91&72.00&80.72\\
            &Llama2 (7B)&88.57&88.06&92.84&63.71&63.32&68.75&76.91&71.36&80.18\\
            &Llama2-Chat (7B)&87.87&87.48&92.92&63.53&63.09&67.79&77.91&71.58&79.79\\
            &Llama2 (13B) &88.64&88.16&93.09&64.08&63.57&69.03&77.43&71.68&80.61\\
            &Llama2-Chat (13B) & 88.59&88.03&92.89&64.32&64.23&68.89&77.89&71.62&80.70\\
		\bottomrule
	\end{tabular}
	\caption{Performance on the validation set of our \detectiondataname. We trained models on the three entailment relations separately.}
    \label{tab:main_eval_validation}
\end{table*}

\end{document}